# Character Spotting using Machine Learning Techniques

Preethi P, Hrishikesh Viswanath
preethip@pes.edu, hrishikeshv@pesu.pes.edu
Department of Computer Science and Engineering, PES University

Abstract This work presents a comparison of machine learning algorithms that are implemented to segment the characters of text presented as an image. The algorithms are designed to work on degraded documents with text that is not aligned in an organized fashion. The paper investigates the use of Support Vector Machines, K-Nearest Neighbor algorithm and an Encoder Network to perform the operation of character spotting. Character Spotting involves extracting potential characters from a stream of text by selecting regions bound by white space. The method works for languages which don't use running script, but rather spaced letters.

Index Terms Neural Networks, Encoder, Character Spotting, Segmentation, Support Vector Machine, K-Nearest Neighbors, image segmentation, machine learning

## I. INTRODUCTION

Interpretation of characters is a necessary requirement in the fields associated with Natural Language Processing [2]. It provides the tools required for understanding arcane languages and reconstructing historical documents. Character spotting refers to the process of identification of potential characters in a stream of text. It doesn't however, include identification or recognition of the characters or even determining whether they are characters of a valid language or merely an illegible pattern.

In order to interpret the message conveyed in such historical documents, it is necessary to extract the characters associated with the text and reconstruct them digitally. Traditional methods of character segmentation [5] fail to meet the required expectations when the documents are inherently noisy. Machine Learning techniques offer ways to handle noisy data and degraded text.

The models are not restricted by the languages and follow a pattern detection approach [14] to determine whether a pattern exists in the window that has been fed to the model.

The text is processed by the models as a sequence of Kernels, containing a portion of the original image. The models try to determine whether the section contains a pattern that is previously encountered by the model during the training process.

## II. RELATED WORK

Text segmentation is the process of extracting words, phrases, topics from sentences. While the work presented in this paper may sound similar to text segmentation, that is not, however the case. The dataset that is used is a collection of digital images, representing estampages. Therefore, the primary objective is image segmentation, where the contents of the image are strings of characters.

In Pal et al.[8], the authors explore the use of various machine learning techniques for image segmentation. Among other models, they discussed the use of neural networks to perform segmentation. They state that the neural networks use contextual data to segment images and are robust to noise. They have considered the problem to be a constraint satisfaction problem. While neural networks are robust to noise, they claim that Networks whose architectures are designed to treat the problem as a CSP are not tested much on noisy data

Guo et al [4] provides a review of semantic segmentation using neural networks. The paper describes the use of RCNN or regions with CNN feature, where the model first detects the objects through search and computes the features of those objects. Classification is later done with support vector machines. The paper also explores another model called FCN, which doesn't require objects to be detected but rather uses CNN layers to perform pixel-to-pixel mapping.

The paper by C Chandhok [1] explores the use of neural networks and K-means clustering to segment images. However, the author shows that K-means clustering works well in homogeneous regions with regards to color and texture.

Zhang et al. [15] propose an architecture to evaluate the quality of segmentation of images using machine learning models. Rather than evaluating the segments manually, they propose the use of another algorithm which evaluates the efficiency of the first one. They have used decision trees which determine the predicted accuracy of that model.

## III. DATASET PREPARATION

The images that are processed by the models are segmented into kernels of fixed dimensions. The data within the kernel is considered complete character if the pattern lies entirely within the kernel. Such a kernel is said to contain a single character. Kernels that contain partially cropped patterns or noise are rejected by the model. The segments are manually labelled for training.

## IV. MODEL DEFINITION AND TRAINING

### A. Data Representation

To classify the images, firstly, the images are represented as an array of features. The feature descriptor that is used is the histogram of oriented gradients [12] feature (HOG), which aids in object detection. The features are block normalized with L2-hys.

To remove unwanted features, PCA is called on the array of HOG values. PCA or Principal Component Analysis [7] is a dimensionality reduction technique that retains uncorrelated data points. PCA is sensitive to scaling and hence, it is necessary to normalize the dataset, which is already done with L2-Hys. PCA is applied on the images, which are taken as matrices of n dimensions. The co-variance matrix is constructed as follows

$$cov(X,Y) = \frac{1}{n-1} \sum_{i=1}^{n} (Xi - \bar{x})(Yi - \bar{y}) \quad (1)$$

The eigen vector matrix of the co-variance matrix is used to transform the data into the new subspace and is done as follows

$$y = W^0 x \quad (2)$$

In the above equation W' is the transpose of the eigen vector matrix of the co-variance matrix cov.

### B. Support Vector Machine

The Support vector machine is a machine learning technique, which when used as a classifier aims to separate the data points into different classes by constructing a hyper plane that separates the elements of the two classes [6]. It first determines the support vectors and further constructs the plane that is at the highest distance from the two vectors [13].

### C. K-Nearest Neighbors

K Nearest neighbors is a technique that classifies a given data element into a category by determining the data points that the given element is most similar to [9]. The category is then determined by the category that the said data points belong to. Conflicts are resolved by determining the category that most of the points fall under. Euclidian distance is used to determine how close two data points are. The distance is calculated as follows

$$d(p,q) = \sqrt{\sum_{i=1}^{n} (q_i - p_i)^2} \quad (3)$$

A K value is chosen to extract K data points that are closest to the point that is to be classified. The output class of the data point is determined by the class that majority of the K points belong to.

A predefined function is used to apply KNN on the HOG feature set on which PCA has been applied. The model is tested against different values of K and the most optimal value is used while assigning a class to a test image.

D. Encoder

The encoder is a special type of a convolutional neural network where the dimensions of the image are reduced at each layer [10]. For an image of dimensions 48x32, six 2D convolution layers are used to reduce the dimensions of the image to 1x1. Each layer is batch normalized and Leaky ReLU is used as the activation function [3]. Two dense layers are added after flattening the output of the convolution layer with Sigmoid as the activation function at the final layer.

The output of the layers may have a large variance. In order to prevent over fitting and improve the rate of learning, the batch of output that is generated by the convolution layers is normalized. The data point x is normalized by subtracting the mean of the batch from the data point and dividing the value obtained by the standard deviation denoted by σ

$$\hat{x} = \frac{x - \mu}{\sqrt{\sigma^2 + \epsilon}} \qquad (4)$$

is required to prevent the denominator from reaching zero. The normalized data point is scaled and shifted.

$$y_i = \gamma \hat{x}_i + \beta \qquad (5)$$

Each hidden layer uses reLU activation function. The convolution layers use padding to prevent the datapoints on the fringe of the matrix to get neglected. Binary Crossentropy (H(p,q)) is calculated between the generated class label and the ground truth [11]. This value, which represents the loss, is minimized to improve the accuracy of the model. Crossentropy is related to entropy, denoted by H(p) and Kullbach Liebler Divergence, which is a measure of how different two distributions are.

The disadvantage of reducing the dimension of the image is loss of data. Since we are not interested in generating the image, we are not worried about calculating MSE and KL Divergence. We are merely extracting information off the image to classify it into two classes.

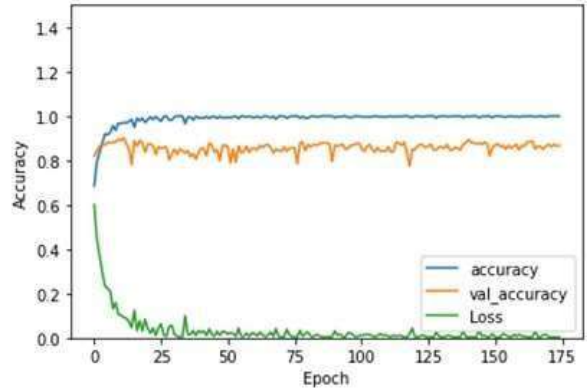

Fig. 1. Training accuracy against epoch

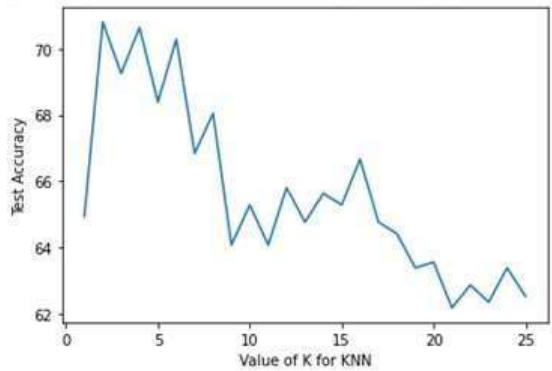

Fig. 2. Variation of accuracy for different values of K

## V. PERFORMANCE AND RESULTS

The following graph represents the training accuracy of the encoder network against epoch.

On comparing the accuracy of the K-Nearest Neighbors model by varying the value of K, it was found that the highest accuracy was achieved when K was set at 3. The accuracy initially fluctuated around 75% and later dropped with the increase in K.

Each of the models performed sub optimally when used individually but the performance improved when the models were cascaded. The wrongly classified output of the first model was re-tested with the second model. The combined accuracy of the two models was found to be significantly higher than when the models were used alone.

| Method Used | Train Accuracy | Train F1 Score | Test Accuracy | Test F1 Score |
|---|---|---|---|---|
| SVM | 63.21% | 67.1% | 53% | 65.6% |
| KNN | 70.81% | 75% | 55% | 58.6.6% |
| Encoder | 99.95% | 99.95% | 58.7% | 33.82% |
| CNN + Encoder | 98.4% | 98.5% | 69.2% | 77% |
| CNN + KNN | 98.4% | 98.5% | 82.53% | 81.9% |

## VI. FUTURE WORK

The dimension of the kernel cannot be dynamically altered since the dimensions of the neural network are directly dependant on the Kernel size. Future work lies in processing characters without resizing to the specified dimension of the network.

## VII. LIMITATIONS

- Severely noisy data mislead the models during training and cause overfitting. The model performs well on images that have low to moderate noise.
- The model cannot segment characters that are connected. Such connected components are categorized as a single compound character.

ACKNOWLEDGEMENT

The work was done while the authors were at PES University. The authors would like to thank Archeological survey of India for providing the dataset.